\documentclass[10pt,twocolumn,letterpaper]{article}

\usepackage{wacv}
\usepackage{times}
\usepackage{epsfig}
\usepackage{graphicx}
\usepackage{amsmath}
\usepackage{amssymb}
\usepackage{booktabs}
\usepackage{caption}
\usepackage{multicol}
\usepackage{subfig}
\usepackage{float}
\usepackage{adjustbox}

\newcommand{\ft}[1]{FaceDancer{#1}} 
\newcommand{\fti}[1]{\textit{FaceDancer}{#1}}  

\def\wacvPaperID{0473} 

\wacvalgorithmstrack   

\wacvfinalcopy 

\ifwacvfinal
\usepackage[breaklinks=true,bookmarks=false]{hyperref}
\else
\usepackage[pagebackref=true,breaklinks=true,colorlinks,bookmarks=false]{hyperref}
\fi

\begin{document}

\title{FaceDancer: Pose- and Occlusion-Aware\\ High Fidelity Face Swapping}

\author{Felix Rosberg$^{1, 2}$
\and
Eren Erdal Aksoy$^2$
\and
Fernando Alonso-Fernandez$^2$
\and
Cristofer Englund$^2$
\and
\small $^1$Berge Consulting, Gothenburg, Sweden
\and
\small $^2$Halmstad University, Halmstad, Sweden
\and
{\tt\small felix.rosberg@berge.io, \{eren.aksoy, fernando.alonso-fernandez, cristofer.englund\}@hh.se}
}



\twocolumn[{%
\renewcommand\twocolumn[1][]{#1}%
\maketitle
\begin{center}
   \centering
    \captionsetup{type=figure}
    \includegraphics[width=0.72\linewidth]{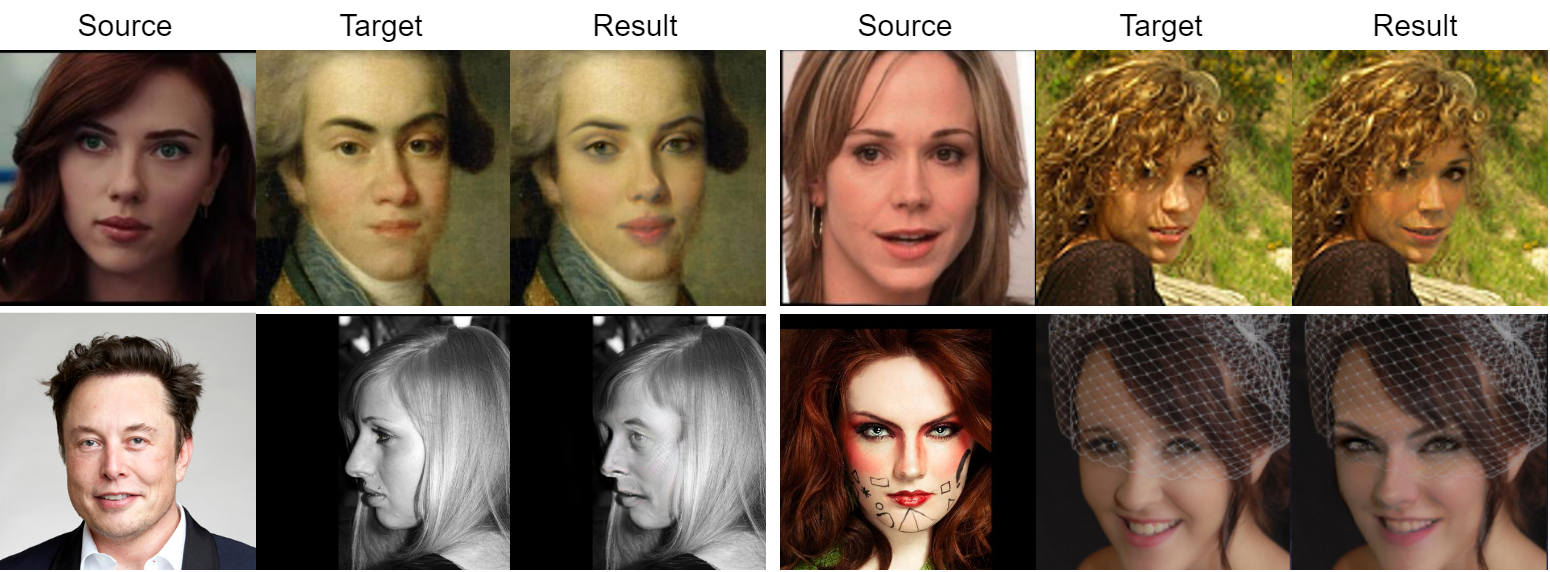}
    \captionof{figure}{Face swapping results generated by \ft.}
    \label{fig:best_results}
\end{center}%
}]

\thispagestyle{empty}


\begin{abstract}
In this work, we present a new  single-stage method  for subject agnostic face swapping and identity transfer, named \fti. 
We have two major contributions: Adaptive Feature Fusion Attention (AFFA) and Interpreted Feature Similarity Regularization (IFSR). 
The AFFA module is embedded in the decoder and adaptively learns to fuse attribute features and features conditioned on identity information without requiring any additional facial segmentation process.  
In IFSR, we leverage  the intermediate features in an identity encoder  to preserve important attributes such as head pose, facial expression, lighting, and occlusion in the target face, while still transferring the identity of the source face with high fidelity.
We conduct extensive quantitative and qualitative experiments on various datasets and show that the proposed \fti~outperforms other state-of-the-art networks in terms of identity transfer, while having significantly better pose preservation than most of the previous  methods. Code available at \url{https://github.com/felixrosberg/FaceDancer}.
\end{abstract}


\section{Introduction}
Face swapping is a challenging task aiming at shifting the identity of a source face into   a target face, while preserving the descriptive face attributes such as facial expression, head pose, and lighting of the target face. The idea of generating such non-existent face pairs has  a vast range of applications in the film, game, and entertainment industry~\cite{3d_emily}.  Therefore, face swapping has rapidly attracted increased research interest in computer vision and graphics.
The   challenge in   swapping faces remains in achieving a high fidelity identity transfer from the source face with a set of attributes which need to be consistent with those in the target face.

There exist two mainstream approaches for face synthesis: \textit{source-oriented} and \textit{target-oriented} methods. 
The former approaches initially synthesize a source face with the  attributes captured in the target face, which is then followed by blending the source face into the target counterpart~\cite{3d_emily},~\cite{fsgan},~\cite{nirkin}. These techniques still have difficulties in handling lighting, occlusion, and complexity. 
The latter approach directly convert   the identity  of the target face into one in the source face~\cite{simswap},~\cite{faceshifter},~\cite{megaface},~\cite{facecontroller},~\cite{hififace},~\cite{ipgan}.
These methods particularly rely on Generative Adversarial Networks (GAN) using a one-stage optimization setting. This helps preserve the target image attributes, such as pose and lighting, without requiring any additional processing step, \eg by learning perceptual and deep features already in the training stage~\cite{simswap},~\cite{faceshifter},~\cite{spade},~\cite{perceptloss1},~\cite{perceptloss2},~\cite{pix2pixhd}.

In this work\footnote{Work done within the Vinnova project MIDAS (2019-05873).},  we introduce a novel, \textit{target-oriented}, and single-stage method, named \fti, to deal  with    challenges, e.g., lighting, occlusion, pose, and semantic structure (See Fig.~\ref{fig:best_results}). \fti~is simple, fast, and accurate. 

\begin{figure*}[!t]
\centering
\includegraphics[width=1.0\textwidth]{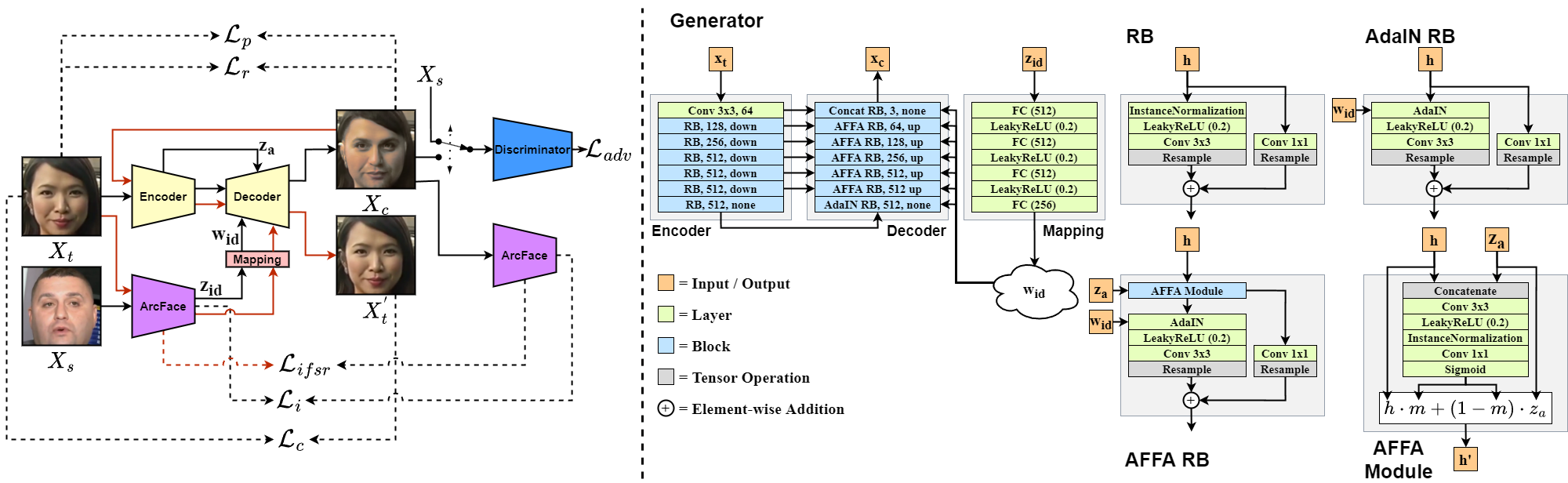}
\caption{Overview of our proposed single-stage face swapping network \fti. Left: The information flow in \fti~during training. Black lines indicate standard information flow, while red lines are the cycle consistency loss information flow and dashed lines represent inputs for losses (Section \ref{sec:losses}). Note that, the ArcFace model has two instances just to avoid having otherwise multiple intersecting arrows in the figure. Right: RB stands for ResBlock. $X_s$ is the source face, $X_t$ is the target face, $X_c$ is the changed face, $z_{id}$ is the identity vector extracted from ArcFace, $w_{id}$ is the mapped identity vector, $h$ is an incoming feature map, and $z_{a}$ is a skip connection feature map. 
The layer Resample represents either an average pooling operation abbreviated as \emph{'down'} or a bilinear upsampling indicated as \emph{'up'} or an identity function shown as \emph{'none'}.
The layer Concat RB concatenates $h$ and $z_a$ without the  AFFA module. }
\label{fig:overview}
\end{figure*}

Our core contribution is twofold: First, we introduce an Adaptive Feature Fusion Attention (AFFA) module, which  adaptively learns during training to produce attention masks that can gate features.
Inspired by the recent methods~\cite{faceshifter} and~\cite{hififace}, the AFFA module is embedded in the decoder and learns attribute features without requiring any additional facial segmentation process.  
The incoming feature maps in AFFA are features that have been conditioned on the source identity information, but also the skip connection of the unconditioned target information in the encoder (See Fig.~\ref{fig:overview}).
The AFFA module, in a nutshell, allows \fti~to learn which conditioned features (\eg identity information) to discard and which unconditioned features (\eg background information) to keep in the target face.
Our experiments show that  gating from the AFFA module considerably improves the identity transfer.

Second, we present an Interpreted Feature Similarity Regularization (IFSR) method  for boosting the attribute preservation.
IFSR regularizes \fti~to  enhance the preservation of facial expression, head pose, and lighting while still transferring the identity with high fidelity. 
More specifically, IFSR explores the similarity between intermediate features in the identity encoder  by comparing the  cosine distance distributions of these features in the target, source, and generated face triplets learned from
a pretrained state-of-the-art identity encoder, ArcFace~\cite{arcface} (See Fig.~\ref{fig:overview}).

We conduct extensive quantitative and qualitative experiments on the FaceForensic++~\cite{faceforensics++} and AFLW2000-3D~\cite{aflw2000} datasets and show that the proposed \fti~significantly outperforms other state-of-the-art networks in terms of identity transfer, while maintaining significantly better pose preservation than most of the previous  methods.
To address the scalability of our network, we further apply \fti~to low resolution images with harsh distortions and qualitatively show that \fti~can still  improve the pose preservation in contrast to other   methods.

\section{Related Work}
There are two leading approaches for face swapping: source- and target-oriented methods. Although our proposed method falls into the later category, we here provide a brief review of the literature related to both approaches. 

\textbf{Source-oriented} approaches first transform the source face to match the expression and posture of the target face, and then blend with the target frame. One of the earliest approaches is The Digital Emily project~\cite{3d_emily} which performs face swapping through expensive and time-consuming 3D scanning of a single actor. Getting one face ready with this method to insert in a scene can, however, take months. Banz et al.~\cite{faceexchange} presents an early approach for utilizing 3D Morphable Models (3DMM) \cite{3dmm} to generate source faces with matching target attributes. This approach, however, comes with the cost that for each image the subject hair must be carefully marked out. 
Nirkin et al.~\cite{nirkin} also utilizes 3DMM to extract pose and expression coefficients from the target face. These coefficients are then employed for reconstructing the source face. The   reconstructed image is finally combined with  the output from a facial segmentation network in order to automate  the entire face swapping process. 
This method, however, struggles with textures and lighting conditions.
FSGAN~\cite{fsgan} introduces a reenactment network particularly designed to reenact the source face based on the target landmarks. In this work, the blending process is performed in an additional step which combines outputs from a segmentation network together with outputs from an inpainting network. This method also struggles with lighting conditions. More importantly, due to relying on the target landmarks for reenactment, the reenacted source falls short in having an effective identity transfer.
\\
\\
\textbf{Target-oriented} approaches mostly rely on generative models to manipulate features of an encoded target face, together with a semi-supervised loss function  or a regularization method to preserve  attributes. Almost all of these methods, including ours, utilize facial recognition models in order to extract identity information to be later used for conditioning of the target features. 
FaceShifter~\cite{faceshifter} robustly transfers the identity  while maintaining  attributes by having an attribute encoder-decoder model trained in a semi-supervised fashion.
This model is coupled with a generator that injects the source identity information and adaptively learns to gate features between the generator and the attribute model.
FaceShifter also has a secondary stage to improve the occlusion awareness. This approach succeeds well with identity and occlusion, but struggles with hard poses, which is solved by our new IFSR loss function. %
SimSwap~\cite{simswap} has an encoder-decoder model that utilizes the identity information to manipulate the bottleneck features.
To preserve attributes, SimSwap uses a modified version of the feature matching loss from pix2pixHD~\cite{pix2pixhd}.
This approach achieves state-of-the-art performance for preserving the pose at an arguably large trade-off for the identity transferability. 
HifiFace~\cite{hififace} uses a combination of GANs and 3DMMs to achieve state-of-the-art identity performance.
Although HifiFace  produces high resolution photo-realistic face swaps, it, however, seems  not to improve the pose considerably and performs  worse than SimSwap. In addition, HifiFace relies on a 3DMM model, which particularly works well with high resolution images only~\cite{3dmmlowres}, \cite{3dmm}. 
 
Our approach differs from these methods in that ours rely on the identity encoder for simplicity and can handle harsh image distortions such as artifacts that emerge in low resolution images. Our method also reaches   state-of-the-art identity performance and improves the pose preservation in contrast to HifiFace.


\section{Method}
\label{s:method}

This section  describes the \fti~network architecture  shown in Fig.~\ref{fig:overview}, together with the AFFA module, the IFSR method, and loss functions. 
Throughout this paper, we use the following notations: $X_t$ refers to the \textit{target face} which is the face image to be manipulated, $X_s$ defines   the \textit{source face} which is the image of the face whose identity is transferred, and $X_c$ is the \textit{changed face} representing the manipulated target face with the identity of the source face.


\subsection{Network Architecture}
\label{ss:netarchitecture}

\fti~involves a generator and a discriminator forming a conditional GAN model coupled with a mapping network and ArcFace \cite{arcface} as depicted in Fig.~\ref{fig:overview}. 

\textbf{Generator:} The generator $G$ relies on an U-Net like encoder-decoder architecture combined with  a mapping network $M$ (See Fig.~\ref{fig:overview}). The encoder consists of a  set of residual blocks with gradually increasing number of filters. The decoder also involves a set of residual blocks, each of which employs either Adaptive Instance Normalization (AdaIN)~\cite{styletransfer},~\cite{stylegan},~\cite{simswap} or an AFFA module or a concatenation layer for exploiting encoded skip connections.
The main aim of $G$ is to generate $X_c$ from the encoded image $X_t$  while 
conditioning the feature maps on the mapped identity vector $w_{id}$  extracted from $X_s$ as shown on the left in Fig.~\ref{fig:overview}. 

\textbf{Discriminator:} The discriminator $D$ used for the adversarial loss is the same as the one in~StarGan-v2 \cite{stargan} and HifiFace~\cite{hififace}, with the exception that we omit the multi-task discrimination, since we use the hinge loss.

\textbf{Mapping network:} \fti~has a mapping network $M$ to boost the performance of $G$ as already shown in~\cite{stylegan},~\cite{stylegan2},~\cite{stylegan3},~\cite{stargan},~\cite{styleganada}. The mapping network learns to transform the initial identity distribution to a new distribution in order to particularly  inject the identity information. The  $M$ network consists of four fully-connected layers (FC) combined with leaky ReLU as non-linearity in all layers except the last (Fig. \ref{fig:overview}).

\textbf{ArcFace:} To extract and inject identity information from the source image $X_s$, \fti~employs a pretrained state-of-the-art identity encoder, ArcFace~\cite{arcface}, coming with  a ResNet50 backbone~\cite{resnet}. 
The resulting ArcFace output is an identity vector with the size of $512$  that serves as an input to \fti.
The ArcFace model is also used for the computation of IFSR (Section~\ref{ss:ifsr}) and the identity loss (Section~\ref{sec:losses}).


\subsection{The Adaptive Feature Fusion Attention (AFFA) Module}
The AFFA module is inspired by previous works such as the Adaptive Attentional Denormalization layer in FaceShifters~\cite{faceshifter}  and  the Semantic Facial Fusion module in HifiFaces~\cite{hififace}.
Unlike the former method where a separate attribute encoder-decoder model exists, we here keep everything condensed within the generator. In contrast to the latter method, which utilizes segmentation masks for supervision,  we here avoid introducing any additional need to compute such segmentation masks for each training sample by letting  AFFA adaptively learn attention masks.
In this regard, AFFA employs the information from skip connections in the generator encoder and forces the generator to learn whether it should rely on features ($z_{a}$) from  skip connections or features ($h$)  from the decoder conditioned on the source identity (Fig. \ref{fig:overview}). This way, AFFA can implicitly learn to extract relevant descriptive face features.
%
Instead of naively concatenating or adding the two feature maps ($h$ and $z_a$),  AFFA first concatenates the feature maps and then passes them through a few learnable layers (Fig. \ref{fig:overview}). 
Finally, AFFA produces an attention mask $m$ with the same filter number as in $h$ and $z_a$. The following equation is used to gate and fuse  $h$ and $z_a$:

\begin{equation}
    h' = h \cdot m + (1 - m) \cdot z_a~~,
\end{equation}

where $h'$ denotes the final fused feature map between $h$ and $z_a$. 
We  experimentally demonstrate the impact of the AFFA module by comparing with cases where either concatenation or addition is individually used to incorporate the information from skip connections in the generator encoder.


\subsection{Interpreted Feature Similarity Regularization (IFSR)}
\label{ss:ifsr}
Target-oriented face swapping methods  rely particularly on semi-supervised or unsupervised techniques  to make sure that the output image maintains the target attributes.
To favor the preservation of attributes, we regularize the \fti~training by employing intermediate features captured by the ArcFace~\cite{arcface}  identity encoder described in section~\ref{ss:netarchitecture}. 
This idea of  using pretrained identity encoders for exploring facial expressions  is also supported by the recent work in~\cite{facenetexp}.

To investigate which layers of ArcFace are responsible for facial expressions and, thus, contribute more to the attribute preservation, we perform a pre-study on a state-of-the-art face swapping model  FaceShifter\cite{faceshifter}. Note that, since the source code of  FaceShifter\cite{faceshifter}, to the best of our knowledge, is not public, we here use our implementation of  FaceShifter with minor modifications. For instance, in our implementation, the generator  down samples  to a resolution of $8\times8$, instead of $2\times2$. We also incorporate the weak feature matching loss from~\cite{simswap} together with the L1 reconstruction loss, instead of L2. 
Next, we use our baseline implementation of FaceShifter to perform random face swaps between identities in the VGGFace2 data set~\cite{vggface2}.
We then compare the cosine distances not only between the target and the generated face swaps, but also between the source  and generated face   pairs for exploring the intermediate features in each block in the ArcFace backbone.
In addition, we compute the distance for intermediate features   between negative pairs (imposters) of identities as qualitative reference. 
All these measured distributions of distances help us determine which layers, i.e., intermediate feature maps, are useful for preserving attribute information. 
For example, if there is a small distance between the target face and the generated face swap, it indicates that the intermediate features from that layer contain  more attribute information.
For this purpose, we also define a margin $m_i$ for each $i$th layer based on the computed mean distances. The motivation for the margin is to regularize the generator to match the mean of the distribution, instead of completely minimizing the distance. The final regularization equation is as follows:

\begin{equation}
    \mathcal{L}_{ifsr} = \sum_{i=k}^{n} min(1 - cos(I^{(i)}(X_t), I^{(i)}(X_c)) - m_{i} \cdot s, 0),
    \label{eq:ifsr}
\end{equation}

where $I^{(i)}$ denotes the  $i$th intermediate feature map in the identity encoder ArcFace, $m_i$ represents the aforementioned margin for the $i$th layer, $s$ is a hyperparameter that scales the margin, $cos(.)$ represents the cosine similarity between two feature maps, $k$ and $n$ respectively denote the index of the first and final blocks, from which intermediate feature maps are extracted. Note that the feature maps are initially reshaped to a vector to have the appropriate dimensionality for the cosine similarity operation. In our experiments, $k$ and $n$ are set to  $2$ and $13$, respectively. 
The main role of the margin scale $s$ is to control the amount of feature similarity that can deviate from the margin.
The lower the value of $s$, the stricter is the similarity.

\subsection{Loss Functions}
\label{sec:losses}
During training, \fti~employs various loss functions: Identity loss, reconstruction loss, perceptual loss, adversarial loss regularized with our IFSR method, and gradient penalty for the discriminator. 
See Fig. \ref{fig:overview} for an overview of how these loss functions interact with inputs and outputs.

The identity loss is used to transfer the source identity as follows:

\begin{equation}
    \mathcal{L}_{i} = 1 - cos(I(X_s), I(X_c))~,
\end{equation}

where $I$ is the identity encoder ArcFace and $cos(.)$ denotes the cosine similarity. The output of $I$ is the identity embedding vector $z_{id}$  (See Fig. \ref{fig:overview}). 

The reconstruction loss is used to make sure that when the target $X_t$ and source $X_s$ are the same images, the final result $X_c$ should be equal to the target image on a pixel-wise level. This reconstruction loss is  defined as follows:

\begin{equation}
  \mathcal{L}_{r}=\begin{cases}
    ||X_t - X_c||, & \text{if} X_t=X_s \\
    0, & \text{otherwise.}
  \end{cases} ~,
\end{equation}

To further strengthen the above behavior and improve the semantic understanding of the image, a perceptual loss is deployed.  The motivation  is that deep features as a perceptual loss have shown to be robust in many reconstruction tasks \cite{pix2pixhd}, \cite{perceptloss1}, \cite{perceptloss2}. The perceptual loss is defined as:

\begin{equation}
  \mathcal{L}_{p}=\begin{cases}
    \sum_{i=0}^{n} ||P^{(i)}(X_t) - P^{(i)}(X_c)||, & \text{if}\ X_t=X_s \\
    0, & \text{otherwise.}
  \end{cases} ~,
\end{equation}

where $P^{(i)}$ denotes the $i$th feature map output of the VGG16 model~\cite{perceptloss1} pretrained on Imagenet~\cite{imagenet} and $n$ is the final index of outputs before the down sample step within the VGG16 model. In our experiments, $n$ is $4$.


Furthermore, we utilize the cycle consistence loss to motivate the model to keep important attributes and structures within the target image~\cite{cyclegan},~\cite{discogan},~\cite{hififace},~\cite{stargan}. The cycle consistence loss is formulated as follows:

\begin{equation}
    \mathcal{L}_{c} = ||X_t - G(X_c, I(X_t))||~,
\end{equation}

where $I$ denotes the identity encoder ArcFace and $G$ is the generator. 

For adversarial loss $\mathcal{L}_{adv}$ we use the hinge loss~\cite{spade},~\cite{spectralnorm},~\cite{sagan},~\cite{biggan},~\cite{funit}. The discriminator is regularized with a gradient penalty term $\mathcal{L}_{gp}$~\cite{wgangp}. The total loss function for the generator $G$ is a weighted sum of above losses, formulated as:

\begin{equation}
    \mathcal{L}_{G} = \mathcal{L}_{adv} + \lambda_{i}\mathcal{L}_{i} + \lambda_{r}\mathcal{L}_{r} + \lambda_{p}\mathcal{L}_{p} + \lambda_{c}\mathcal{L}_{c} + \lambda_{ifsr}\mathcal{L}_{ifsr}~,
\end{equation}

where $\lambda_{i} = 10$, $\lambda_{r} = 5$, $\lambda_{p} = 0.2$, $\lambda_{c} = 1$ and $\lambda_{ifsr} = 1$. The weighting for $\mathcal{L}_{gp}$ ($\lambda_{gp}$) is set to 10.

\section{Results}
\textbf{Implementation Details:} \fti~is trained on the datasets VGGFace2~\cite{vggface2} and LS3D-W~\cite{ls3dw}. All faces are aligned with five point landmarks extracted with RetinaFace~\cite{retinaface}. The alignment is performed to match the input into ArcFace~\cite{arcface}. We keep all images in the data sets. ArcFace is pretrained on MS1M~\cite{ms1m} with a ResNet50 backbone. We used the Adam~\cite{adam} optimizer with $\beta_1 = 0$, $\beta_2 = 0.99$, a learning rate of 0.0001, and exponential learning rate decay of $0.97$ every 100K steps. The target ($X_t$) and source ($X_s$) images are randomly augmented with brightness, contrast, and saturation.
Each configuration is trained for 300K steps for the ablation study (Table~\ref{t:ablations} and Table~\ref{t:compare2}). 
We further train   all the best performing configurations in the ablation studies (B, C, D) up to 500K steps to compare with the recent works using a batch size of 10.
Image resolution for all of our models are $256\times256$. There is a $20\%$ chance that an image pair is the same, with at least one pair in the batch being the same. Margin scale $s$ in Eq.~\ref{eq:ifsr} is set to $1.2$.


\subsection{Quantitative Results}
\label{s:quant}

\begin{table}[t!]
\caption{Quantitative experiments on FaceForensics++~\cite{faceforensics++}. See Table~\ref{t:ablations} for the definition of each \fti~configuration (Config B to D). These models has been trained for 500k iterations.}
\label{t:compare}
\begin{center}
\scalebox{0.95}{
\begin{tabular}{p{3.1cm}cccc}
\hline
Method & ID$\uparrow$ & Pose$\downarrow$ & Exp$\downarrow$ & FID$\downarrow$\\
\hline
FaceSwap~\cite{faceswap} & 54.19 & 2.51 & N/A & N/A\\
FaceShifter~\cite{faceshifter} & 97.38 & 2.96 & N/A & N/A\\
MegaFS~\cite{megaface} & 90.83 & 2.64 & N/A & N/A \\
FaceController~\cite{facecontroller} & 98.27 & 2.65 & N/A & N/A \\
HifiFace~\cite{hififace} & 98.48 & 2.63 & N/A & N/A\\
SimSwap~\cite{simswap} & 92.83 & \textbf{1.53} & 8.04 & \textbf{11.76}\\
\hline
FaceDancer ({\scriptsize Config B}) & 98.54 & 2.24 & 8.52 & 25.11\\
FaceDancer ({\scriptsize Config C}) & \textbf{98.84} & 2.04 & 7.97 & 16.30\\
FaceDancer ({\scriptsize Config D}) & 98.19 & 2.15 & \textbf{5.70} & 19.10\\
\hline
\end{tabular}
}
\end{center}
\end{table}

\begin{figure*}[!t]
\centering
\includegraphics[width=0.9\textwidth]{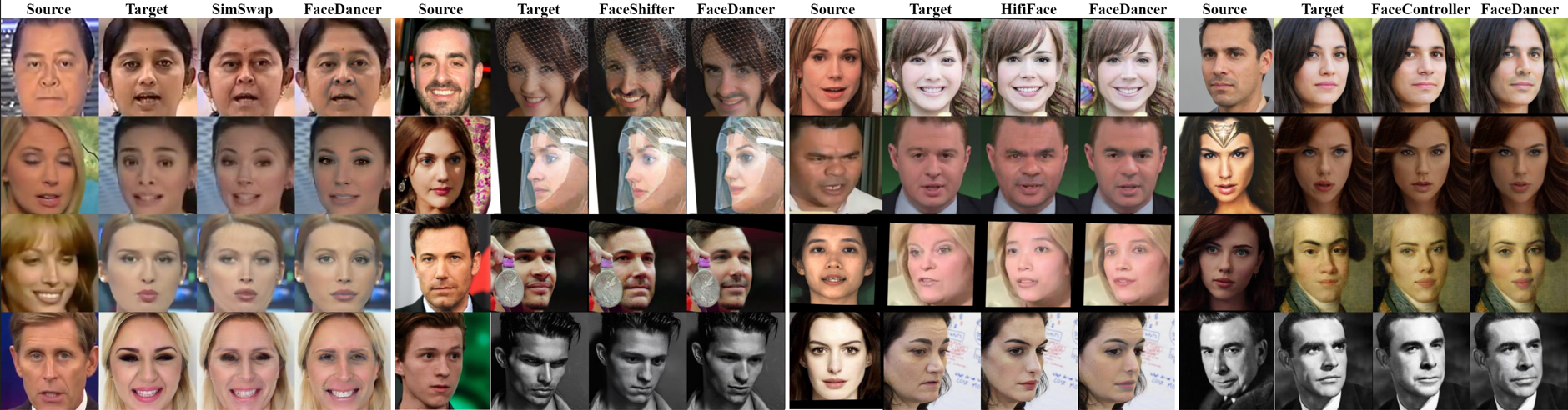}
\caption{Comparing our model \ft~ with SimSwap~\cite{simswap}, FaceShifter~\cite{faceshifter}, HifiFace~\cite{hififace}, and FaceController~\cite{facecontroller}.}
\label{fig:compall}
\end{figure*}

We perform quantitative evaluation of  \fti~using the FaceForensics++~\cite{faceforensics++} dataset and compare it to the other state-of-the-art face swapping networks, such as  SimSwap~\cite{simswap}, FaceShifter~\cite{faceshifter}, HifiFace~\cite{hififace}, and FaceController~\cite{facecontroller}. 
The metrics evaluated are identity retrieval (ID), pose error, expression error, and Frechét Inception Distance (FID)~\cite{fid}. 
For the identity retrieval, we initially perform random swaps for each image in the test set and then retrieve the correct identity with a secondary identity encoder, CosFace~\cite{cosface}. 
To compare pose, we use the  pose estimator in~\cite{pose}  and report the average L2 error. The expression metric is often omitted for comparison due to poor accessibility of models. However, we here use an implementation of an expression embedder~\cite{expression} and report the average L2 error. 
FID is calculated between the swapped version of the test set and the unaltered test set and helps demonstrate when a model has problems with lighting, occlusion, visual quality, and posture.

Similar to the previous works~\cite{faceshifter},~\cite{simswap},~\cite{hififace}, we sample 10 frames from each video in FaceForensic++ which yields a test data set of 10K. 
As shown in Table~\ref{t:compare} our method \fti~outperforms all the previous works by leading to the highest  identity retrieval performance. Regarding the pose metric, we have comparable results, i.e., \fti~achieves  the second-lowest pose error ($2.04$) after SimSwap~\cite{simswap}.

\subsection{Qualitative Results}

For the qualitative evaluation,  we compare the performance of our model \fti~with the recent state-of-the-art works SimSwap~\cite{simswap}, FaceShifter~\cite{faceshifter}, HifiFace~\cite{hififace}, and FaceController~\cite{facecontroller} as shown in Fig. \ref{fig:compall}.
We here note that SimSwap~\cite{simswap} is the only work coming with a public and easy to access model. Due to this fact,  we   have more in depth comparison with SimSwap, whereas for the other baseline models we show qualitative results for sample images only reported in these works. 

Fig.~\ref{fig:compall} shows that our model \fti~behaves similar to  SimSwap, but one can easily notice the substantially improved identity transfer in our results. 
FaceShifter performs good identity transfer and is able to transfer relevant attributes such as facial hair while preserving occlusion and the identity face shape. 
FaceShifter, however, struggles with lighting and gaze direction as it heavily relies on the second stage model. FaceController exhibits good identity transferability and decent pose error, however, still fails noticeably often with the gaze direction. 
Our approach \fti~deals with all these problems better. Finally, HifiFace demonstrates promising results regarding all these metrics, particularly when it comes to the facial shape. For instance, HifiFace exhibits better face shape preservation of the identity than our model. Otherwise, it is not feasible to compare qualitatively with HifiFace since our model \fti~quantitatively performs better (See Table~\ref{t:compare}).

\begin{figure}[!t]
\centering
\includegraphics[width=6.5cm]{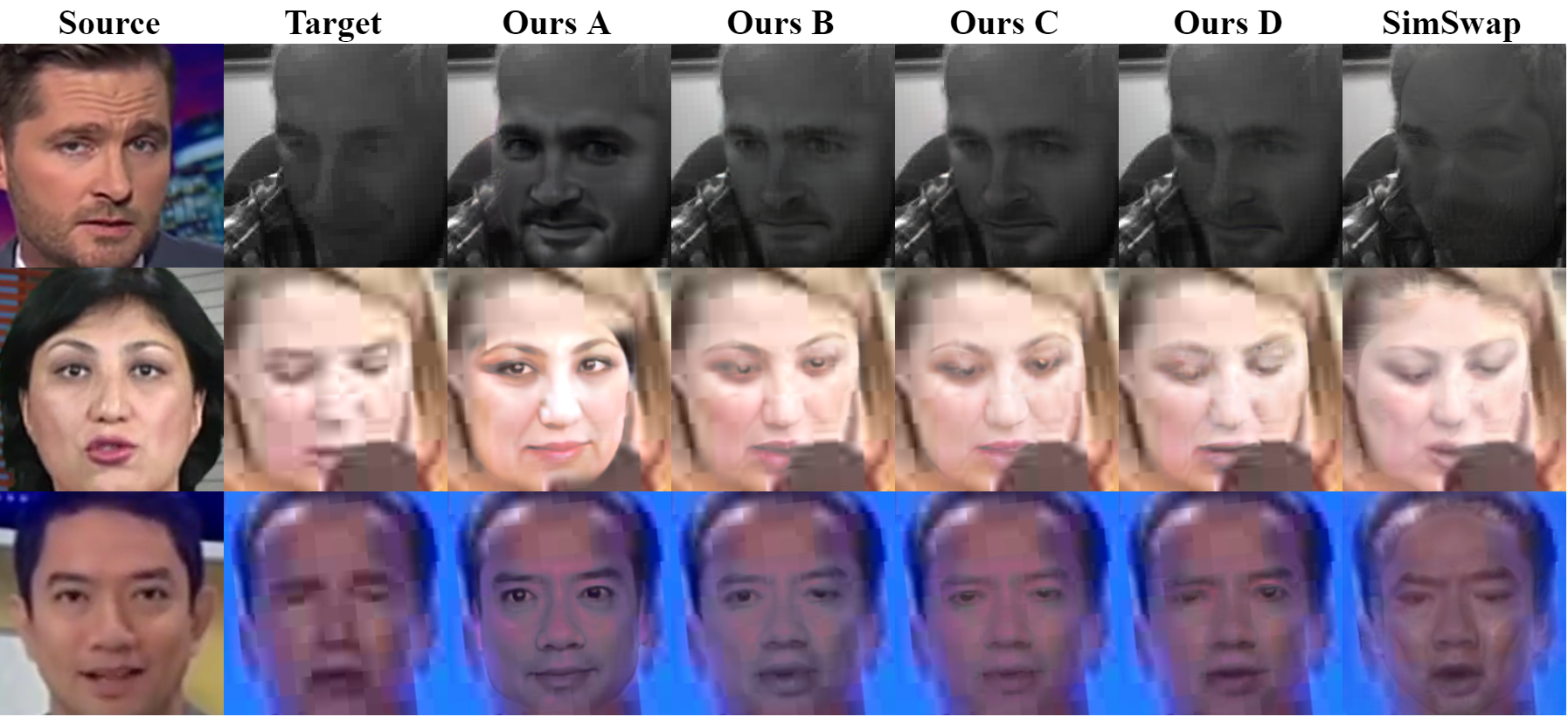}
\caption{Qualitative comparison on low resolution images. See Table~\ref{t:ablations} for the definition of each \fti~configuration (Config B to D).}
\label{fig:lowres}
\end{figure}

Furthermore, to address the scalability of our model, we qualitatively analyze the performance of \fti~compared to SimSwap on low resolution face images. 
Fig.~\ref{fig:lowres} shows that \fti~has enough capacity to capture the semantic structure of the face images  even under low resolution cases. 
\fti~is able to maintain the pixelation artifacts, while SimSwap either produces a smooth face or completely fails, as depicted in the first row of Fig.~\ref{fig:lowres}. \fti~also works well on videos without any temporal information. We refer to supplementary materials for video results. In supplementary materials we also include further results of images in higher resolution, further comparisons, occlusion, difficult poses, extreme cases and finally failure cases. Failure usually occurs when the face poses away from the camera or the face pose is an uncommon angle represented in the data.

\subsection{Ablation Study}

\begin{figure}[!t]
\centering
\includegraphics[width=6.0cm]{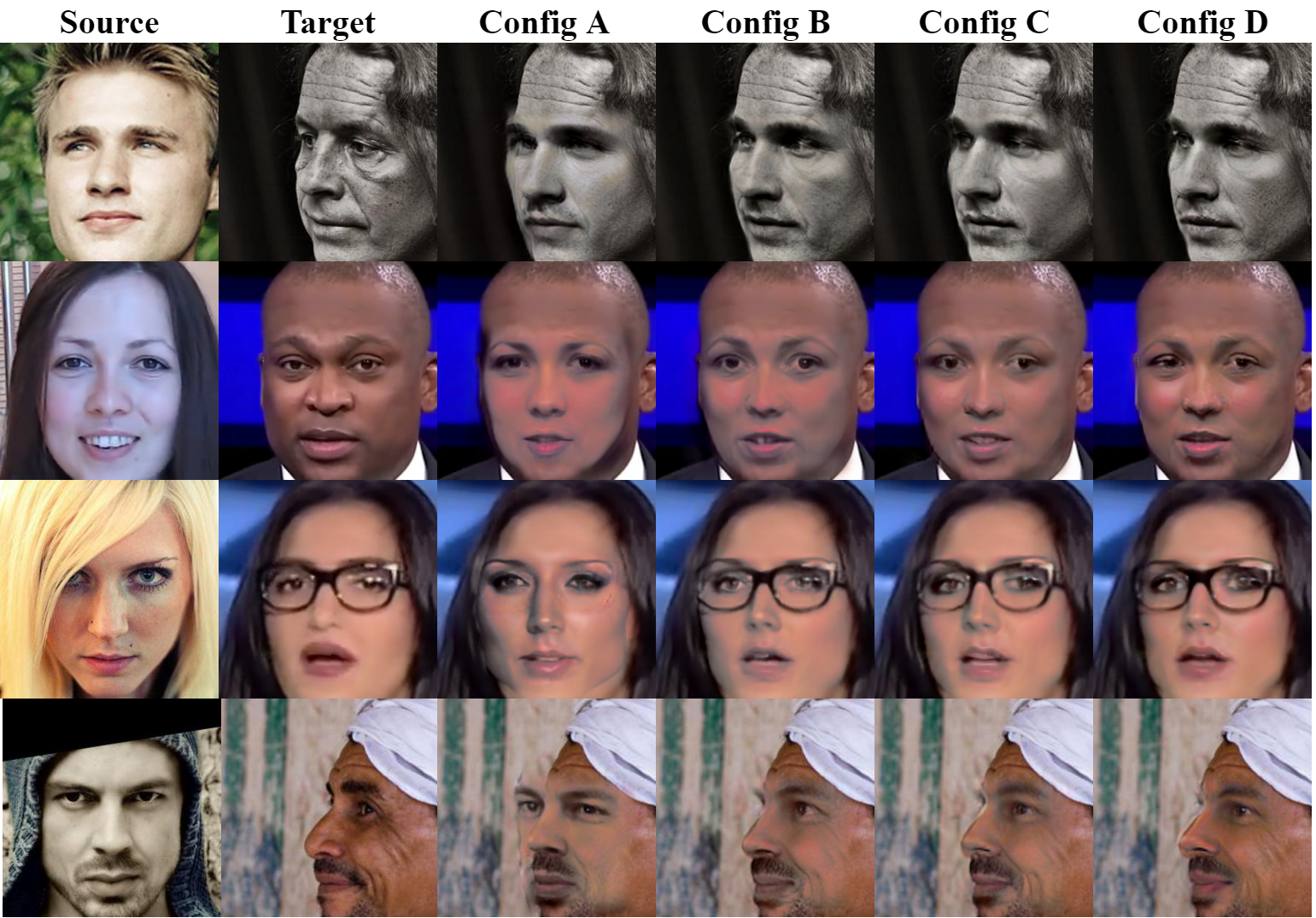}
\caption{Illustration of the impact of IFSR. Config A given in the 3rd column here shows results once IFSR is omitted during  training as described in Table \ref{t:ablations}.}
\label{fig:configcomp}
\end{figure}



\begin{table*}[!t]
\centering
\caption{Ablative analysis together with the runtime performance.  Inference time is given in millisecond and memory usage in GB. All models in this table were trained for 300k iterations.}
\label{t:ablations}
\begin{center}
\scalebox{0.85}{
\begin{tabular}{c|c|c|c|c|c|cccc|cc}
\hline
Config & IFSR & AFFA & Concat final skip* & 6 skips & Mapping & ID$\uparrow$ & Pose$\downarrow$ & Exp$\downarrow$ & FID$\downarrow$ & Inference & Memory\\
\hline
{\scriptsize Baseline 1} & \checkmark & - & - & - & \checkmark & 97.66 & 1.97 & 8.20 & 16.72 & 74.9 & 1.25\\
{\scriptsize Baseline 2} & \checkmark & - & - & - & \checkmark & 92.61 & \textbf{1.87} & 7.97 & 13.51 & 70.2  & 1.25 \\
A & - & \checkmark & - & - & \checkmark & 98.14 & 3.61 & 9.82 & 31.63 & 75.8 & \textbf{1.18} \\
B & \checkmark & \checkmark & - & - & \checkmark & 96.96 & 2.48 & 8.25 & 23.11 & 75.8  & \textbf{1.18} \\
C & \checkmark & \checkmark & \checkmark & - & \checkmark & \textbf{98.57} & 2.27 & 7.98 & 14.59 & 78.3  & 1.26 \\
D & \checkmark & \checkmark & \checkmark & \checkmark & \checkmark & 97.53 & 2.04 & 7.76 & \textbf{13.50} & 78.2  & 1.27 \\
E & \checkmark & \checkmark & \checkmark & \checkmark & - & 97.38 & 2.07 & \textbf{5.73} & 14.68 & \textbf{64.6}  & 1.21 \\

\hline
\end{tabular}
}
\end{center}
{\scriptsize * Concatenation instead of AFFA at resolution 256 + one extra AFFA modules at resolution 32. See supplementary materials for detailed figures for each configuration.}
\end{table*}

We here ablate different \fti~components (such as the AFFA module and the IFSR method) and compare to two baselines as shown in Table~\ref{t:ablations}. The ablations shown in Table~\ref{t:ablations} are evaluated on FaceForensic++~\cite{faceforensics++} and the ablations shown in Table~\ref{t:compare2} are evaluated on AFLW2000-3D~\cite{aflw2000}.
Baseline 1 and 2 respectively employ concatenation and addition in order to fuse feature maps from the decoder and skip connection. 
For baseline 1, baseline 2, configuration A and configuration B, the feature fusion is performed at top three resolutions (256, 128, 64).
For configuration C, we use concatenation at resolution 256 and AFFA at resolution 128, 64, and 32 are processed.
Configuration D is the same as C, but with two additional AFFA modules and skips at resolution 16 and 8 (Fig. \ref{fig:overview}). 
Configuration E is the same as D with the only difference that the mapping network ($M$) in the generator is omitted (Fig. \ref{fig:overview}). We refer to the supplementary materials for detailed figures of the baselines and configurations.
All ablation  configurations are trained for 300K steps.
As reported in Table~\ref{t:ablations}, baselines 1 and 2 achieve the lowest pose errors, however, with the cost of having either high FID score or poor identity performance.
Configuration A improves identity performance but does not use IFSR which leads to poor pose error, expression error, and FID. Since Configuration B employs IFSR, it improves the expression and pose problem, however, still struggles with the FID.
Configuration C overcomes these problems and achieves state-of-the-art performance on identity. 
Adding two more AFFA modules in lower resolutions in the decoder slightly disrupts the identity performance, but improves the other metrics further. This is mainly because Configuration D fuses more features from the target face. The last row in this table shows that the mapping information employed by the \fti~generator improves identity transfer and FID with expression error as trade off.
 
In Table~\ref{t:ablations}, we also provide the total runtime performance for each \fti~configuration. Inference and memory consumption profiling are done on a single Nvidia RTX 3090 with a batch size of 32. Profiling includes inference for ArcFace.

The contribution of IFSR and AFFA becomes clearer when we ablate on the pose challenging dataset AFLW2000-3D~\cite{aflw2000} (Table~\ref{t:compare2}). We here use AFLW2000-3D as the target data set and FaceForsenic++ as the source data set.
In this case, after randomly swapping all faces in AFLW2000-3D with faces from FaceForensic++, we try to retrieve the original identity in FaceForensic++. 

Our findings in Table~\ref{t:compare2} depict that baseline 1 still performs the best for pose, but falls short on the other metrics.
Configurations A through D perform significantly better for ID, however, they have comparable pose error and similar or better expression error. Configuration E  demonstrates  the impact of not having the mapping network $M$ used in \fti. Configuration E falls short for identity performance and pose error (Table~\ref{t:compare2}).

\begin{table}[!t]
\caption{Ablative analysis using AFLW2000-3D~\cite{aflw2000} as target and FaceForensics++~\cite{faceforensics++} as source. See Table~\ref{t:ablations} for configuration details.}
\label{t:compare2}
\begin{center}
\begin{tabular}{c|cccc}
\hline
Config & ID$\uparrow$ & Pose$\downarrow$ & Exp$\downarrow$ & FID$\downarrow$\\
\hline
{\scriptsize Baseline 1} & 89.10 & \textbf{5.63} & 5.34 & 19.26\\
{\scriptsize Baseline 2} & 94.95 & 6.23 & 5.60 & 21.30\\
A & \textbf{98.50} & 14.97 & 7.07 & 40.34\\
B & 97.95 & 5.86 & 5.74 & 21.50\\
C & 97.65 & 5.82 & \textbf{4.13} & 18.50\\
D & 97.10 & 5.75 & 4.15 & 20.41\\
E & 95.45 & 6.16 & 4.19 & \textbf{18.13}\\

\hline
\end{tabular}
\end{center}
\end{table}

\begin{figure*}[!t]
\centering
\subfloat{\centering{{\includegraphics[width=5.5cm]{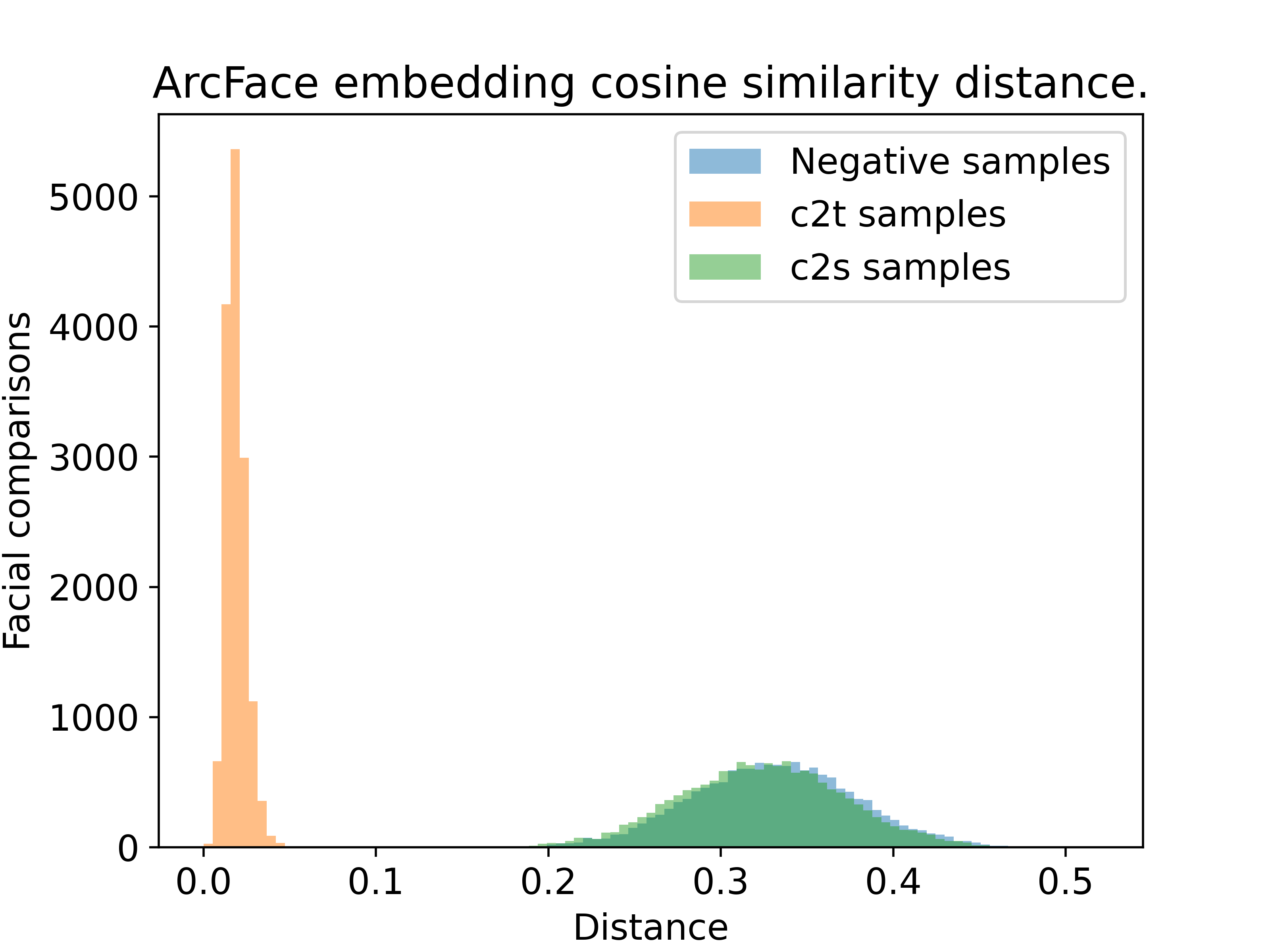}}}}
\subfloat{\centering{{\includegraphics[width=5.5cm]{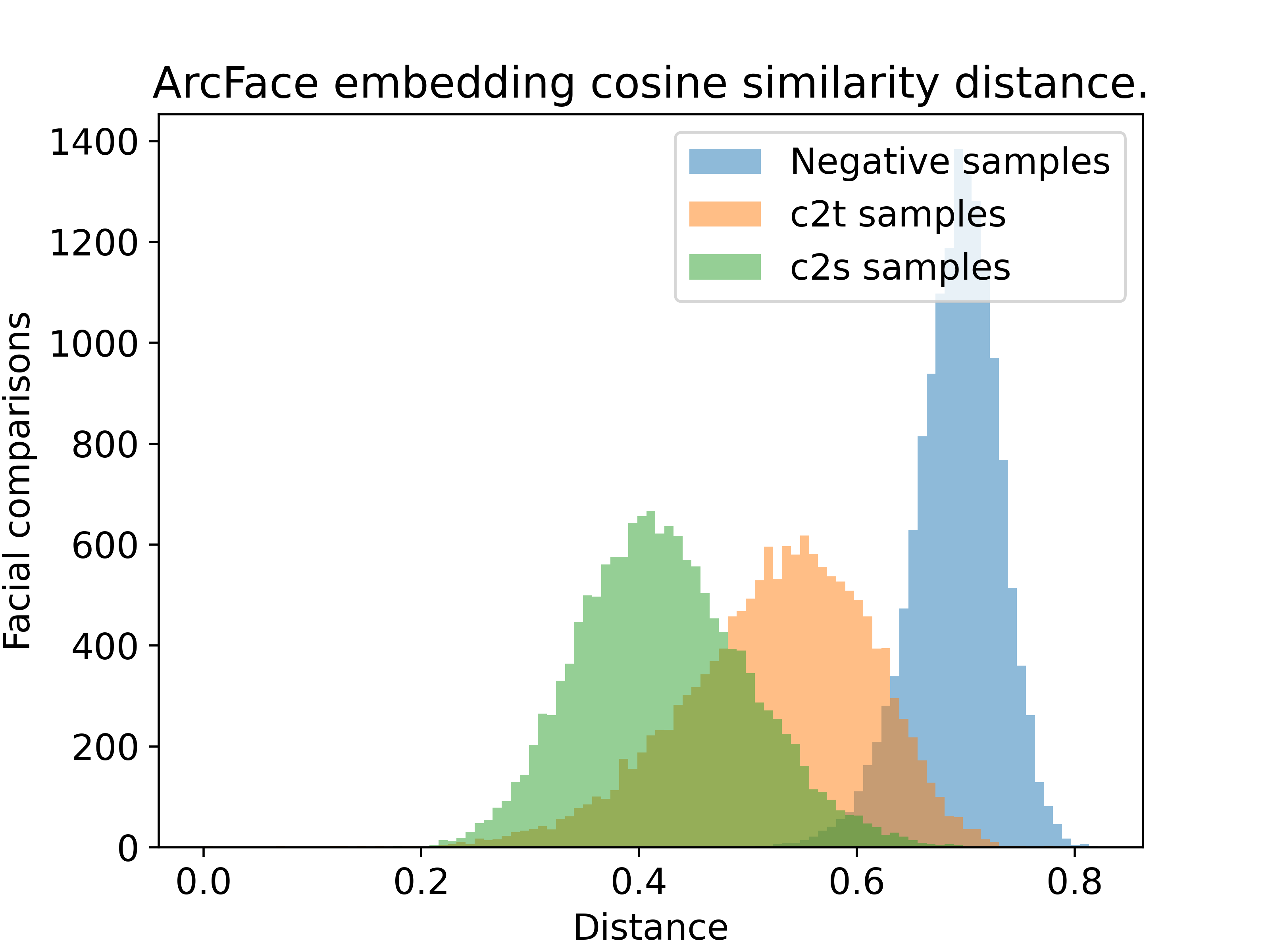}}}}
\subfloat{\centering{{\includegraphics[width=5.5cm]{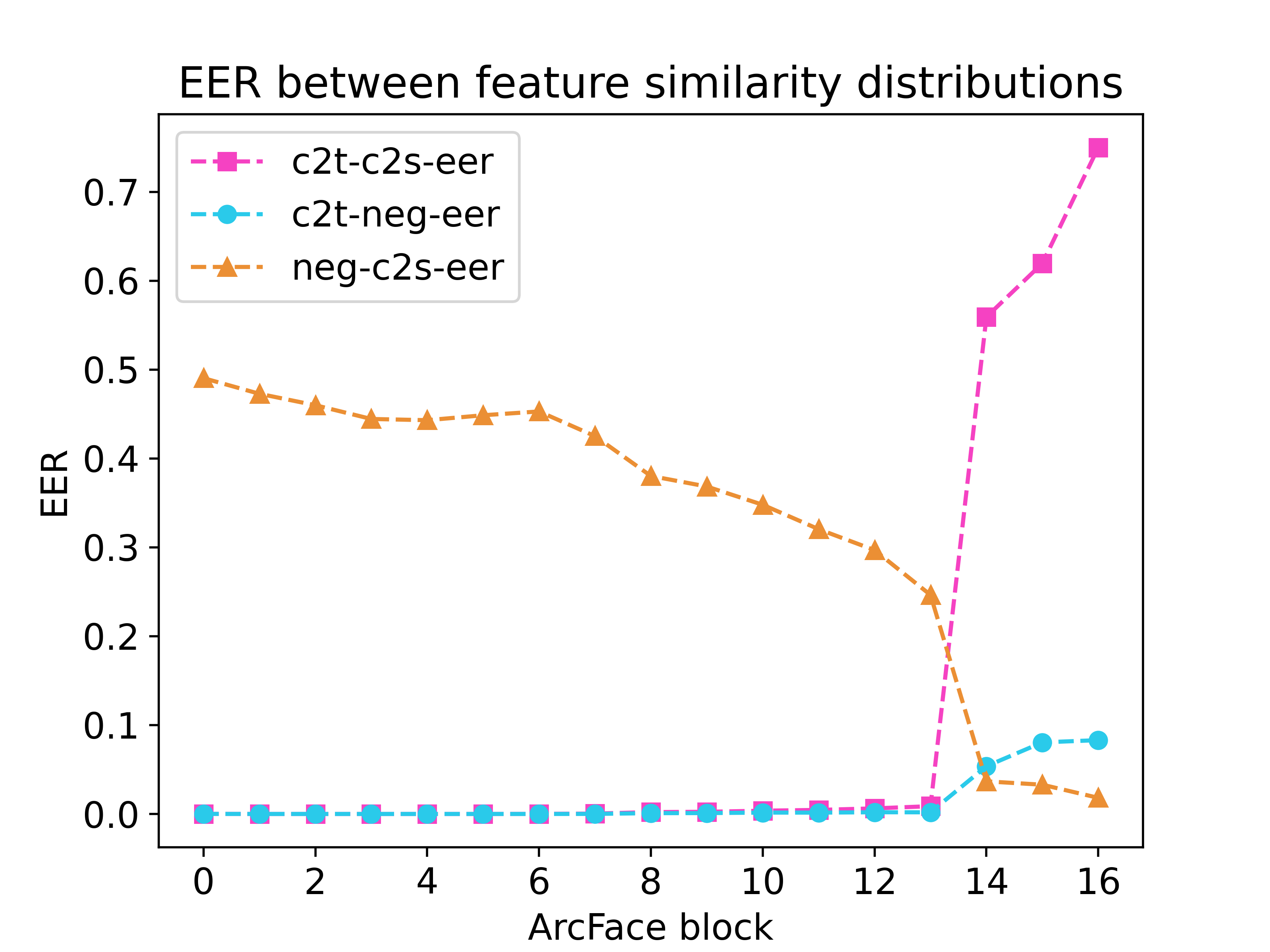}}}}
\caption{Cosine similarity between intermediate features between changed and target faces (c2t), changed and source faces (c2s), and different identities (Negative Samples).
(a) Distances between features from first block of ArcFace. 
(b) Distances between features from final block of ArcFace.
(c) Equal error rates (EER) between the distance distributions for intermediate features in every block.}
\label{fig:blocks}
\end{figure*}

\begin{figure}
\centering {{\includegraphics[width=8cm]{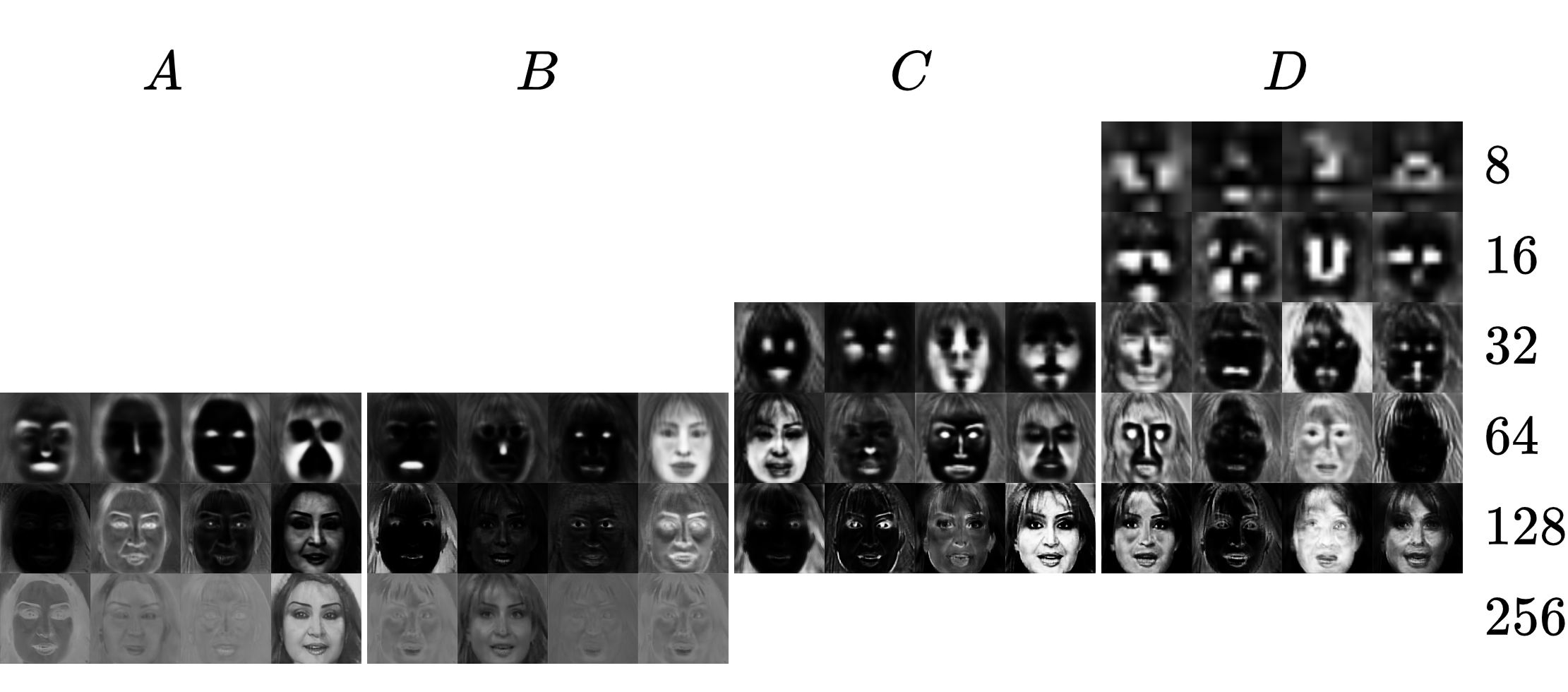}}}
\caption{Comparison between example attention maps at different resolutions for different configurations in Table~~\ref{t:ablations}.}
\label{fig:atterr}
\end{figure}
\subsection{Analysis of the AFFA Module}

In this section, we provide a comprehensive study showing the role of the   Adaptive Feature Fusion Attention (AFFA) module. 
For this purpose, we first  trained \fti~using AFFA in the three upper resolutions of the decoder (256, 128, 64). However, this leads to noticeable color defects in the swapped face images (Fig.~\ref{fig:configcomp}). 
Experimental findings reported in Fig.~\ref{fig:atterr} show that this is due to the usage of AFFA in the end of \fti~generator. 
The attention maps generated at the resolution of 256 are mostly gray, as depicted in Fig.~\ref{fig:atterr}.
We hypothesize that in the highest layer of the generator, the attention maps are far away from fusing the feature maps as expected.
As shown in Table~\ref{t:ablations}, baselines 1 and 2  do not have any problems with the color defect as demonstrated by the substantially lower FID scores compared to configurations A and B which  rely on AFFA at resolution 256. 
To remedy this problem, we replace the final AFFA module with a simple concatenation operation while adding an AFFA module to resolution 32.
In Fig.~\ref{fig:atterr}, we show examples of attentions maps for each configuration in Table~\ref{t:ablations} at each resolution of the decoder where the \fti~generator uses an AFFA module.

\subsection{Analysis of IFSR}

We now provide a comprehensive experimental evaluation on the role of the  Interpreted Feature Similarity Regularization  (IFSR) method. 

We start investigating intermediate features within the ArcFace ResNet50 backbone by comparing the cosine distance between feature maps computed for the target face, the source face, the changed face, and negative pairs using the VGGFace2~\cite{vggface2} dataset. This process is repeated for each residual block output in ArcFace.
The changed face is obtained by deploying a pretrained implementation of FaceShifter~\cite{faceshifter}, briefly detailed in Section~\ref{ss:ifsr}.
As shown in Fig.~\ref{fig:blocks}(a), the changed and target faces ($c2t$) share significantly more similar features than those observed between the changed and source faces ($c2s$) in the early ArcFace layers. 
This behavior is not observed in the final residual block, as depicted in Fig.~\ref{fig:blocks}(b). 
This strongly suggests that the identity encoder contains important information such as pose, expression and occlusions in the earlier layers while the final blocks store identity information.
To demonstrate the separability of the $c2t$ and $c2s$ distributions in Fig.~\ref{fig:blocks}, we calculate the equal error rate (EER) between these distributions. 
As shown by the EER plots in Fig.~\ref{fig:blocks}(c), the $c2t$ and $c2s$ distributions are completely separable until block 14. 
Afterwards, the EER jumps to more than 50\%, which means that the $c2t$ distribution moves to the right of $c2s$, i.e., $X_c$ shares more identity attributes with $X_s$ in contrast to $X_t$ in the same layer. This confirms that $X_c$ successfully captures the identity of $X_s$.
The qualitative impact of our proposed IFSR method is shown in Fig.~\ref{fig:configcomp}. Without IFSR, the pasted face effect and lack of expression preservation become more apparent.
Note that the layers and information from IFSR come from a frozen identity encoder. Therefore, any pretrained face swap framework could here be employed to calculate the IFSR margins.  IFSR itself does not contain any learnable parameter. The process is  just needed to gain an interpretable insight of what kind of information the layers contain (expression, pose, color, lightning, identity, etc.), and how to define the margins for IFSR.


\section{Conclusion}
In this work, we introduce \fti~as a new single-stage  face swapping model that quantitatively reaches state-of-the-art. 
\fti~has a novel  regularization component IFSR which utilizes intermediate features to preserve attributes such as pose, facial expression, and occlusion. 
Furthermore, the AFFA module in \fti~drastically improves identity transfer without a significant trade off for visual quality and attribute preservation when coupled with IFSR. \fti~is limited in two main aspects, transferring face shape and the need of calculating IFSR margins from a pretrained face swap model. Future directions for the latter is to figure out how to calculate the margins adaptively online. IFSR can potentially be used to compress complex face swap (or even image translation) models. Trying to combine IFSR with 3DMM to gain strong pose, occlusion and face shape preservation would be interesting future work.

{\small
\bibliographystyle{ieee_fullname}
\bibliography{egbib}

\begin{thebibliography}{10}\itemsep=-1pt

\bibitem{faceswap}
Faceswap.
\newblock Accessed 2022-02-18.

\bibitem{3d_emily}
Oleg Alexander, Mike Rogers, William Lambeth, Matt Chiang, and Paul Debevec.
\newblock Creating a photoreal digital actor: The digital emily project.
\newblock In {\em 2009 Conference for Visual Media Production}, pages 176--187,
  2009.

\bibitem{ipgan}
Jianmin Bao, Dong Chen, Fang Wen, Houqiang Li, and Gang Hua.
\newblock Towards open-set identity preserving face synthesis.
\newblock In {\em 2018 IEEE/CVF Conference on Computer Vision and Pattern
  Recognition}, pages 6713--6722, 2018.

\bibitem{faceexchange}
Volker Blanz, Kristina Scherbaum, Thomas Vetter, and Hans-Peter Seidel.
\newblock Exchanging faces in images.
\newblock In {\em Computer Graphics Forum}, volume~23, pages 669--676. Wiley
  Online Library, 2004.

\bibitem{biggan}
Andrew Brock, Jeff Donahue, and Karen Simonyan.
\newblock Large scale {GAN} training for high fidelity natural image synthesis.
\newblock In {\em 7th International Conference on Learning Representations,
  {ICLR} 2019, New Orleans, LA, USA, May 6-9, 2019}. OpenReview.net, 2019.

\bibitem{ls3dw}
Adrian Bulat and Georgios Tzimiropoulos.
\newblock How far are we from solving the 2d \& 3d face alignment problem?(and
  a dataset of 230,000 3d facial landmarks).
\newblock In {\em Proceedings of the IEEE International Conference on Computer
  Vision}, pages 1021--1030, 2017.

\bibitem{vggface2}
Qiong Cao, Li Shen, Weidi Xie, Omkar~M Parkhi, and Andrew Zisserman.
\newblock Vggface2: A dataset for recognising faces across pose and age.
\newblock In {\em 2018 13th IEEE international conference on automatic face \&
  gesture recognition (FG 2018)}, pages 67--74. IEEE, 2018.

\bibitem{simswap}
Renwang Chen, Xuanhong Chen, Bingbing Ni, and Yanhao Ge.
\newblock {\em SimSwap: An Efficient Framework For High Fidelity Face
  Swapping}, page 2003–2011.
\newblock Association for Computing Machinery, New York, NY, USA, 2020.

\bibitem{stargan}
Yunjey Choi, Youngjung Uh, Jaejun Yoo, and Jung-Woo Ha.
\newblock Stargan v2: Diverse image synthesis for multiple domains.
\newblock In {\em Proceedings of the IEEE/CVF conference on computer vision and
  pattern recognition}, pages 8188--8197, 2020.

\bibitem{imagenet}
Jia Deng, Wei Dong, Richard Socher, Li-Jia Li, Kai Li, and Li Fei-Fei.
\newblock Imagenet: A large-scale hierarchical image database.
\newblock In {\em 2009 IEEE Conference on Computer Vision and Pattern
  Recognition}, pages 248--255, 2009.

\bibitem{retinaface}
Jiankang Deng, Jia Guo, Evangelos Ververas, Irene Kotsia, and Stefanos
  Zafeiriou.
\newblock Retinaface: Single-shot multi-level face localisation in the wild.
\newblock In {\em Proceedings of the IEEE/CVF Conference on Computer Vision and
  Pattern Recognition}, pages 5203--5212, 2020.

\bibitem{arcface}
Jiankang Deng, Jia Guo, Niannan Xue, and Stefanos Zafeiriou.
\newblock Arcface: Additive angular margin loss for deep face recognition.
\newblock In {\em 2019 IEEE/CVF Conference on Computer Vision and Pattern
  Recognition (CVPR)}, pages 4685--4694, 2019.

\bibitem{3dmm}
Bernhard Egger, William~AP Smith, Ayush Tewari, Stefanie Wuhrer, Michael
  Zollhoefer, Thabo Beeler, Florian Bernard, Timo Bolkart, Adam Kortylewski,
  Sami Romdhani, et~al.
\newblock 3d morphable face models—past, present, and future.
\newblock {\em ACM Transactions on Graphics (TOG)}, 39(5):1--38, 2020.

\bibitem{wgangp}
Ishaan Gulrajani, Faruk Ahmed, Martin Arjovsky, Vincent Dumoulin, and Aaron~C
  Courville.
\newblock Improved training of wasserstein gans.
\newblock {\em Advances in neural information processing systems}, 30, 2017.

\bibitem{ms1m}
Yandong Guo, Lei Zhang, Yuxiao Hu, Xiaodong He, and Jianfeng Gao.
\newblock Ms-celeb-1m: A dataset and benchmark for large-scale face
  recognition.
\newblock In {\em European conference on computer vision}, pages 87--102.
  Springer, 2016.

\bibitem{resnet}
Kaiming He, Xiangyu Zhang, Shaoqing Ren, and Jian Sun.
\newblock Deep residual learning for image recognition.
\newblock In {\em Proceedings of the IEEE conference on computer vision and
  pattern recognition}, pages 770--778, 2016.

\bibitem{fid}
Martin Heusel, Hubert Ramsauer, Thomas Unterthiner, Bernhard Nessler, and Sepp
  Hochreiter.
\newblock Gans trained by a two time-scale update rule converge to a local nash
  equilibrium.
\newblock {\em Advances in neural information processing systems}, 30, 2017.

\bibitem{3dmmlowres}
Guosheng Hu, Chi~Ho Chan, Josef Kittler, and Bill Christmas.
\newblock Resolution-aware 3d morphable model.
\newblock In {\em BMVC}, pages 1--10. University of Surrey, 2012.

\bibitem{styletransfer}
Xun Huang and Serge Belongie.
\newblock Arbitrary style transfer in real-time with adaptive instance
  normalization.
\newblock In {\em Proceedings of the IEEE international conference on computer
  vision}, pages 1501--1510, 2017.

\bibitem{perceptloss1}
Justin Johnson, Alexandre Alahi, and Li Fei-Fei.
\newblock Perceptual losses for real-time style transfer and super-resolution.
\newblock In Bastian Leibe, Jiri Matas, Nicu Sebe, and Max Welling, editors,
  {\em Computer Vision -- ECCV 2016}, pages 694--711, Cham, 2016. Springer
  International Publishing.

\bibitem{styleganada}
Tero Karras, Miika Aittala, Janne Hellsten, Samuli Laine, Jaakko Lehtinen, and
  Timo Aila.
\newblock Training generative adversarial networks with limited data.
\newblock {\em Advances in Neural Information Processing Systems},
  33:12104--12114, 2020.

\bibitem{stylegan3}
Tero Karras, Miika Aittala, Samuli Laine, Erik H{\"a}rk{\"o}nen, Janne
  Hellsten, Jaakko Lehtinen, and Timo Aila.
\newblock Alias-free generative adversarial networks.
\newblock {\em Advances in Neural Information Processing Systems}, 34, 2021.

\bibitem{stylegan}
Tero Karras, Samuli Laine, and Timo Aila.
\newblock A style-based generator architecture for generative adversarial
  networks.
\newblock In {\em Proceedings of the IEEE/CVF conference on computer vision and
  pattern recognition}, pages 4401--4410, 2019.

\bibitem{stylegan2}
Tero Karras, Samuli Laine, Miika Aittala, Janne Hellsten, Jaakko Lehtinen, and
  Timo Aila.
\newblock Analyzing and improving the image quality of stylegan.
\newblock In {\em Proceedings of the IEEE/CVF conference on computer vision and
  pattern recognition}, pages 8110--8119, 2020.

\bibitem{discogan}
Taeksoo Kim, Moonsu Cha, Hyunsoo Kim, Jung~Kwon Lee, and Jiwon Kim.
\newblock Learning to discover cross-domain relations with generative
  adversarial networks.
\newblock In {\em International conference on machine learning}, pages
  1857--1865. PMLR, 2017.

\bibitem{adam}
Diederik~P Kingma and Jimmy Ba.
\newblock Adam: A method for stochastic optimization.
\newblock {\em arXiv preprint arXiv:1412.6980}, 2014.

\bibitem{faceshifter}
Lingzhi Li, Jianmin Bao, Hao Yang, Dong Chen, and Fang Wen.
\newblock Advancing high fidelity identity swapping for forgery detection.
\newblock In {\em Proceedings of the IEEE/CVF Conference on Computer Vision and
  Pattern Recognition}, pages 5074--5083, 2020.

\bibitem{funit}
Ming-Yu Liu, Xun Huang, Arun Mallya, Tero Karras, Timo Aila, Jaakko Lehtinen,
  and Jan Kautz.
\newblock Few-shot unsupervised image-to-image translation.
\newblock In {\em Proceedings of the IEEE/CVF International Conference on
  Computer Vision}, pages 10551--10560, 2019.

\bibitem{spectralnorm}
Takeru Miyato, Toshiki Kataoka, Masanori Koyama, and Yuichi Yoshida.
\newblock Spectral normalization for generative adversarial networks.
\newblock In {\em International Conference on Learning Representations}, 2018.

\bibitem{fsgan}
Yuval Nirkin, Yosi Keller, and Tal Hassner.
\newblock Fsgan: Subject agnostic face swapping and reenactment.
\newblock In {\em 2019 IEEE/CVF International Conference on Computer Vision
  (ICCV)}, pages 7183--7192, 2019.

\bibitem{nirkin}
Yuval Nirkin, Iacopo Masi, Anh Tran~Tuan, Tal Hassner, and Gerard Medioni.
\newblock On face segmentation, face swapping, and face perception.
\newblock In {\em 2018 13th IEEE International Conference on Automatic Face
  Gesture Recognition (FG 2018)}, pages 98--105, 2018.

\bibitem{spade}
Taesung Park, Ming-Yu Liu, Ting-Chun Wang, and Jun-Yan Zhu.
\newblock Semantic image synthesis with spatially-adaptive normalization.
\newblock In {\em Proceedings of the IEEE Conference on Computer Vision and
  Pattern Recognition}, 2019.

\bibitem{expression}
Felix Rosberg and Cristofer Englund.
\newblock Comparing facial expressions for face swapping evaluation with
  supervised contrastive representation learning.
\newblock In {\em 2021 16th IEEE International Conference on Automatic Face and
  Gesture Recognition (FG 2021)}, pages 01--05, 2021.

\bibitem{faceforensics++}
Andreas Rossler, Davide Cozzolino, Luisa Verdoliva, Christian Riess, Justus
  Thies, and Matthias Nie{\ss}ner.
\newblock Faceforensics++: Learning to detect manipulated facial images.
\newblock In {\em Proceedings of the IEEE/CVF International Conference on
  Computer Vision}, pages 1--11, 2019.

\bibitem{pose}
Nataniel Ruiz, Eunji Chong, and James~M Rehg.
\newblock Fine-grained head pose estimation without keypoints.
\newblock In {\em Proceedings of the IEEE conference on computer vision and
  pattern recognition workshops}, pages 2074--2083, 2018.

\bibitem{facenetexp}
Raviteja Vemulapalli and Aseem Agarwala.
\newblock A compact embedding for facial expression similarity.
\newblock In {\em Proceedings of the IEEE/CVF Conference on Computer Vision and
  Pattern Recognition}, pages 5683--5692, 2019.

\bibitem{cosface}
Hao Wang, Yitong Wang, Zheng Zhou, Xing Ji, Dihong Gong, Jingchao Zhou, Zhifeng
  Li, and Wei Liu.
\newblock Cosface: Large margin cosine loss for deep face recognition.
\newblock In {\em Proceedings of the IEEE conference on computer vision and
  pattern recognition}, pages 5265--5274, 2018.

\bibitem{pix2pixhd}
Ting-Chun Wang, Ming-Yu Liu, Jun-Yan Zhu, Andrew Tao, Jan Kautz, and Bryan
  Catanzaro.
\newblock High-resolution image synthesis and semantic manipulation with
  conditional gans.
\newblock In {\em Proceedings of the IEEE Conference on Computer Vision and
  Pattern Recognition}, 2018.

\bibitem{hififace}
Yuhan Wang, Xu Chen, Junwei Zhu, Wenqing Chu, Ying Tai, Chengjie Wang, Jilin
  Li, Yongjian Wu, Feiyue Huang, and Rongrong Ji.
\newblock Hififace: 3d shape and semantic prior guided high fidelity face
  swapping.
\newblock In Zhi-Hua Zhou, editor, {\em Proceedings of the Thirtieth
  International Joint Conference on Artificial Intelligence, {IJCAI-21}}, pages
  1136--1142. International Joint Conferences on Artificial Intelligence
  Organization, 8 2021.
\newblock Main Track.

\bibitem{facecontroller}
Zhiliang Xu, Xiyu Yu, Zhibin Hong, Zhen Zhu, Junyu Han, Jingtuo Liu, Errui
  Ding, and Xiang Bai.
\newblock Facecontroller: Controllable attribute editing for face in the wild.
\newblock {\em Proceedings of the AAAI Conference on Artificial Intelligence},
  35(4):3083--3091, May 2021.

\bibitem{sagan}
Han Zhang, Ian Goodfellow, Dimitris Metaxas, and Augustus Odena.
\newblock Self-attention generative adversarial networks.
\newblock In {\em International conference on machine learning}, pages
  7354--7363. PMLR, 2019.

\bibitem{perceptloss2}
Richard Zhang, Phillip Isola, Alexei~A Efros, Eli Shechtman, and Oliver Wang.
\newblock The unreasonable effectiveness of deep features as a perceptual
  metric.
\newblock In {\em Proceedings of the IEEE conference on computer vision and
  pattern recognition}, pages 586--595, 2018.

\bibitem{cyclegan}
Jun-Yan Zhu, Taesung Park, Phillip Isola, and Alexei~A Efros.
\newblock Unpaired image-to-image translation using cycle-consistent
  adversarial networks.
\newblock In {\em Proceedings of the IEEE international conference on computer
  vision}, pages 2223--2232, 2017.

\bibitem{aflw2000}
Xiangyu Zhu, Zhen Lei, Xiaoming Liu, Hailin Shi, and Stan~Z Li.
\newblock Face alignment across large poses: A 3d solution.
\newblock In {\em Proceedings of the IEEE conference on computer vision and
  pattern recognition}, pages 146--155, 2016.

\bibitem{megaface}
Yuhao Zhu, Qi Li, Jian Wang, Cheng-Zhong Xu, and Zhenan Sun.
\newblock One shot face swapping on megapixels.
\newblock In {\em Proceedings of the IEEE/CVF Conference on Computer Vision and
  Pattern Recognition (CVPR)}, pages 4834--4844, June 2021.

\end{thebibliography}
}

\newpage

\onecolumn
\section*{Supplementary Material}
In this section we show further results, including edge cases and failure cases. You can also find video results here\footnote{\url{https://drive.google.com/drive/folders/1hHjK0W-Oo1HD6OZb97IdSifPs4_c6NNo?usp=sharing}} or at the github page. You will also find network structures of the remaining configurations that was tested in the ablations study below.

\begin{adjustbox}{center,caption={Face swap matrix results from \fti . When source and target is the same, the result visually appears to be perfect reconstruction.},label={resultmatrix},nofloat=figure}
\includegraphics[width=1.0\textwidth]{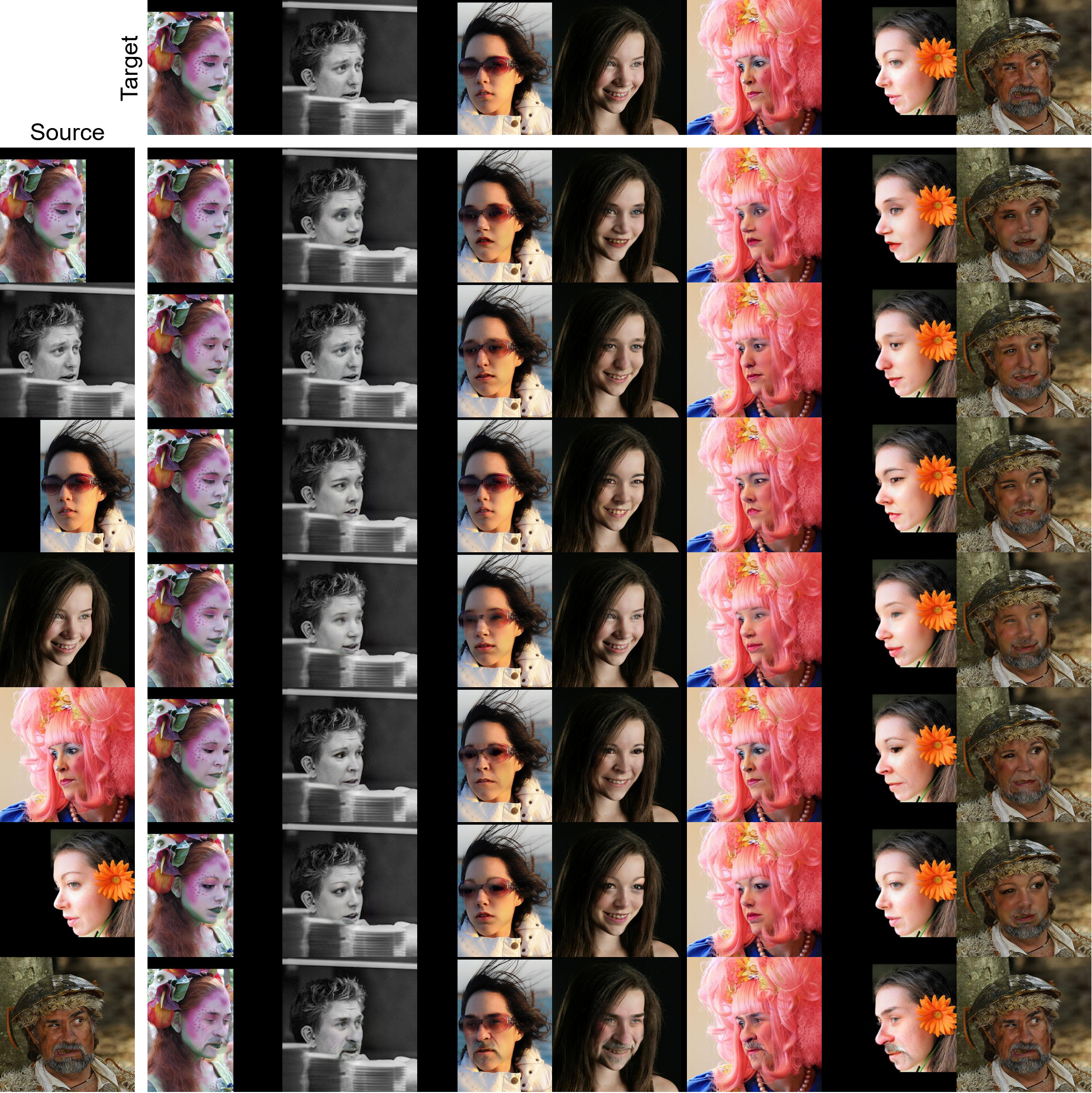}
\end{adjustbox}

\begin{adjustbox}{center,caption={Further results from FaceDancer and comparison with SimSwap. The left column display images with challenging occlusion or lightning. The right column display images with challenging facial poses.},label={fig:examples},nofloat=figure}
\includegraphics[width=1.0\textwidth]{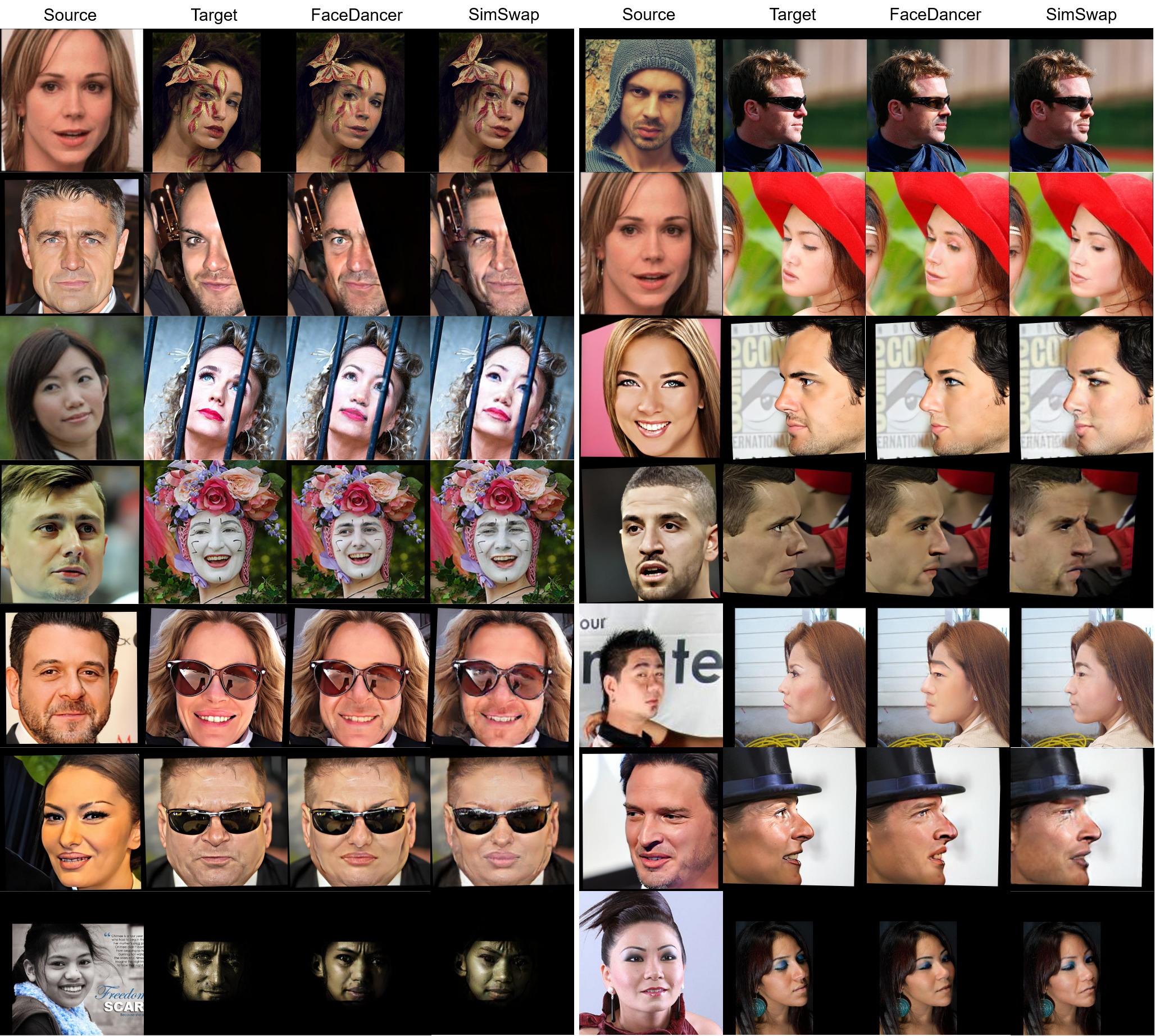}
\end{adjustbox}

\begin{adjustbox}{center,caption={Further results in extreme and difficult cases and comparisons with SimSwap. The left column display face swaps with challenging images, such as extreme occlusions and expressions. The right column display failure cases. Most failure cases occur when the face is posing away from the camera, but can also occur in rare lighting conditions.},label={fig:failures},nofloat=figure}
\includegraphics[width=1.0\textwidth]{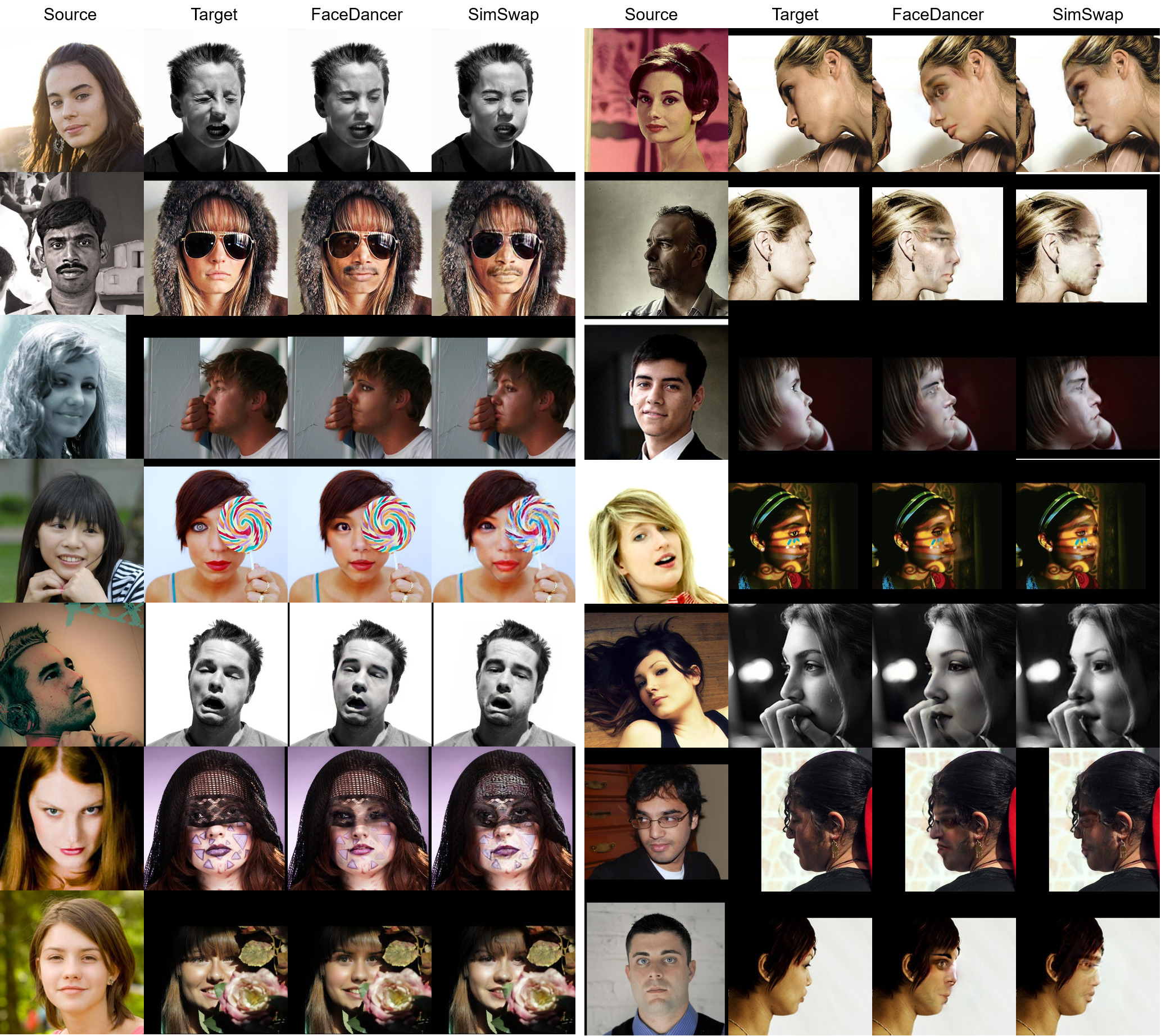}
\end{adjustbox}

\begin{figure*}[htp]
\centering
\includegraphics[width=0.42\textwidth]{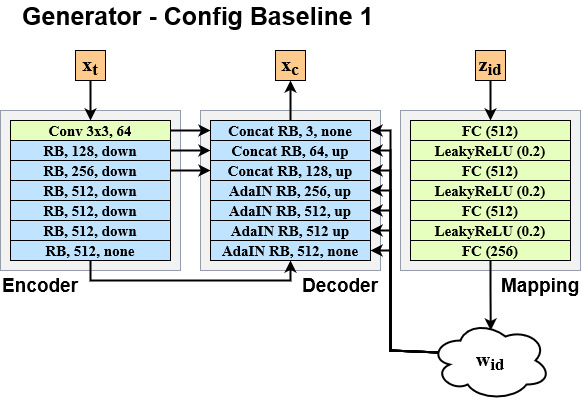} ~~~~
\includegraphics[width=0.42\textwidth]{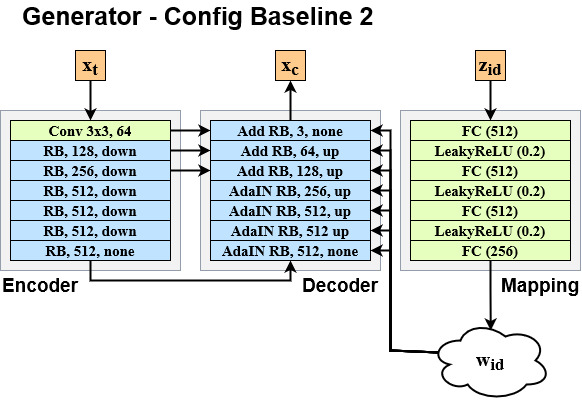}
\caption{Overview of baseline 1 and baseline 2 of the \fti.}
\label{fig:baseline1}
\end{figure*}

\begin{figure*}[htp]
\centering
\includegraphics[width=0.42\textwidth]{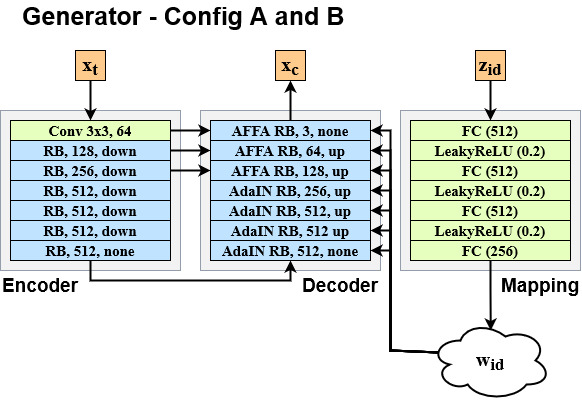} ~~~~
\includegraphics[width=0.42\textwidth]{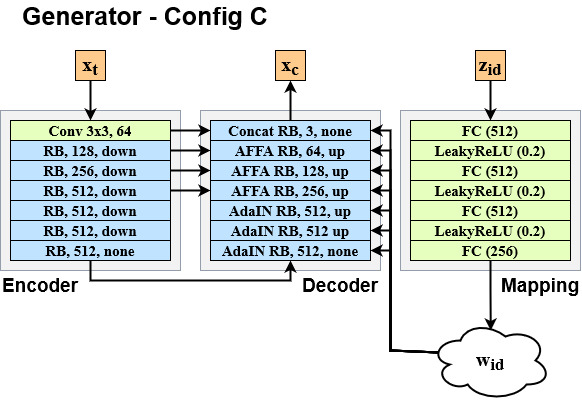}
\caption{Overview of configuration A, B (the same structure) and C of the \fti. Note that the difference between the Configs A and B is the IFSR module, which is not included in the Config A. }
\label{fig:configabc}
\end{figure*}

\begin{figure*}[htp]
\centering
\includegraphics[width=0.37\textwidth]{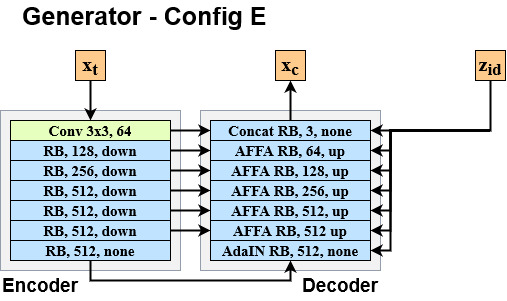} ~~~~
\includegraphics[width=0.57\textwidth]{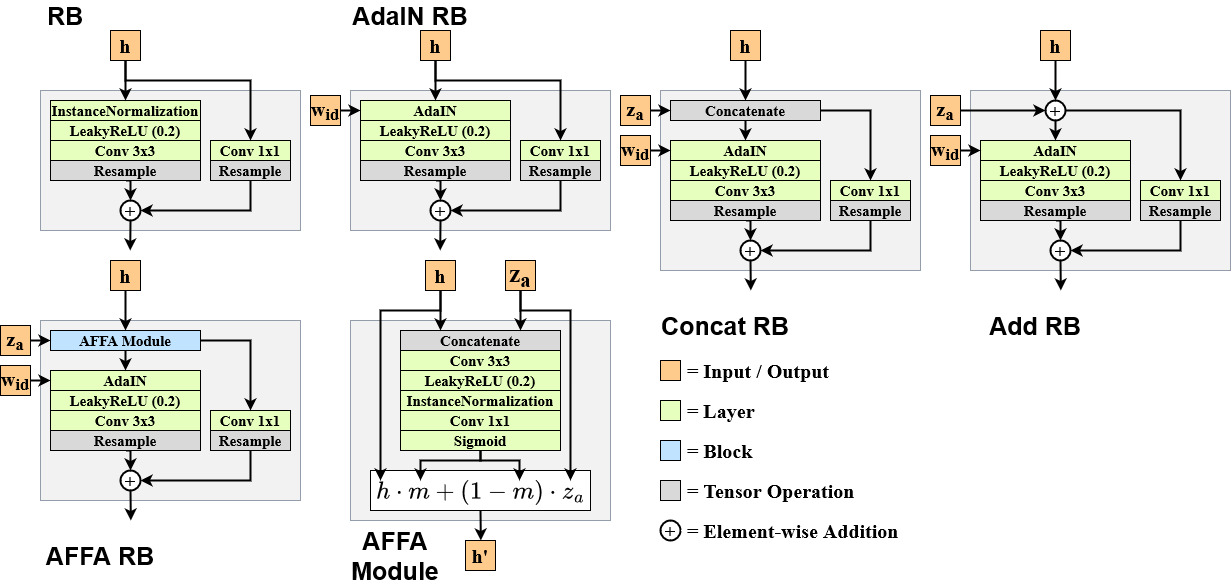}
\caption{Overview of configuration E of the \fti~ and block details.}
\label{fig:confige}
\end{figure*}

\end{document}